\documentclass[letterpaper, 10pt, conference]{ieeeconf}

\IEEEoverridecommandlockouts
\usepackage[T1]{fontenc}
\usepackage[utf8]{inputenc}
\usepackage{graphicx}
\usepackage{booktabs}
\usepackage{amsmath,amssymb}
\usepackage{url}
\usepackage{xspace}

\usepackage[style=numeric, sorting=none]{biblatex}
\usepackage{hyperref}
\hypersetup{hypertex=true,
colorlinks=true,
linkcolor=blue,
anchorcolor=blue,
citecolor=blue,
urlcolor=cyan   
}

\newcommand{\gsw}{Genie Sim PanoRecon\xspace}
\newcommand{\gs}{Genie Sim\xspace}

\title{\LARGE \bf
Genie Sim PanoRecon: Fast Immersive Scene Generation \\ from Single-View Panorama
}

\author{%
Zhijun Li$^{\dagger}$,
Yongxin Su$^{\dagger}$,
Di Yang$^{\dagger}$,
Jichao Wang$^{\ddagger}$,
Zheyuan Xing$^{\ddagger}$,
Qian Wang$^{*}$,
Maoqing Yao$^{*}$%
\thanks{$^{\dagger}$Equal contribution (first authors). $^{\ddagger}$Equal contribution (second authors).}%
\thanks{$^{*}$Co-corresponding authors}%
}
\IEEEaftertitletext{ 
    \centering 
    \vspace{10pt} 
    \includegraphics[width=0.9\linewidth]{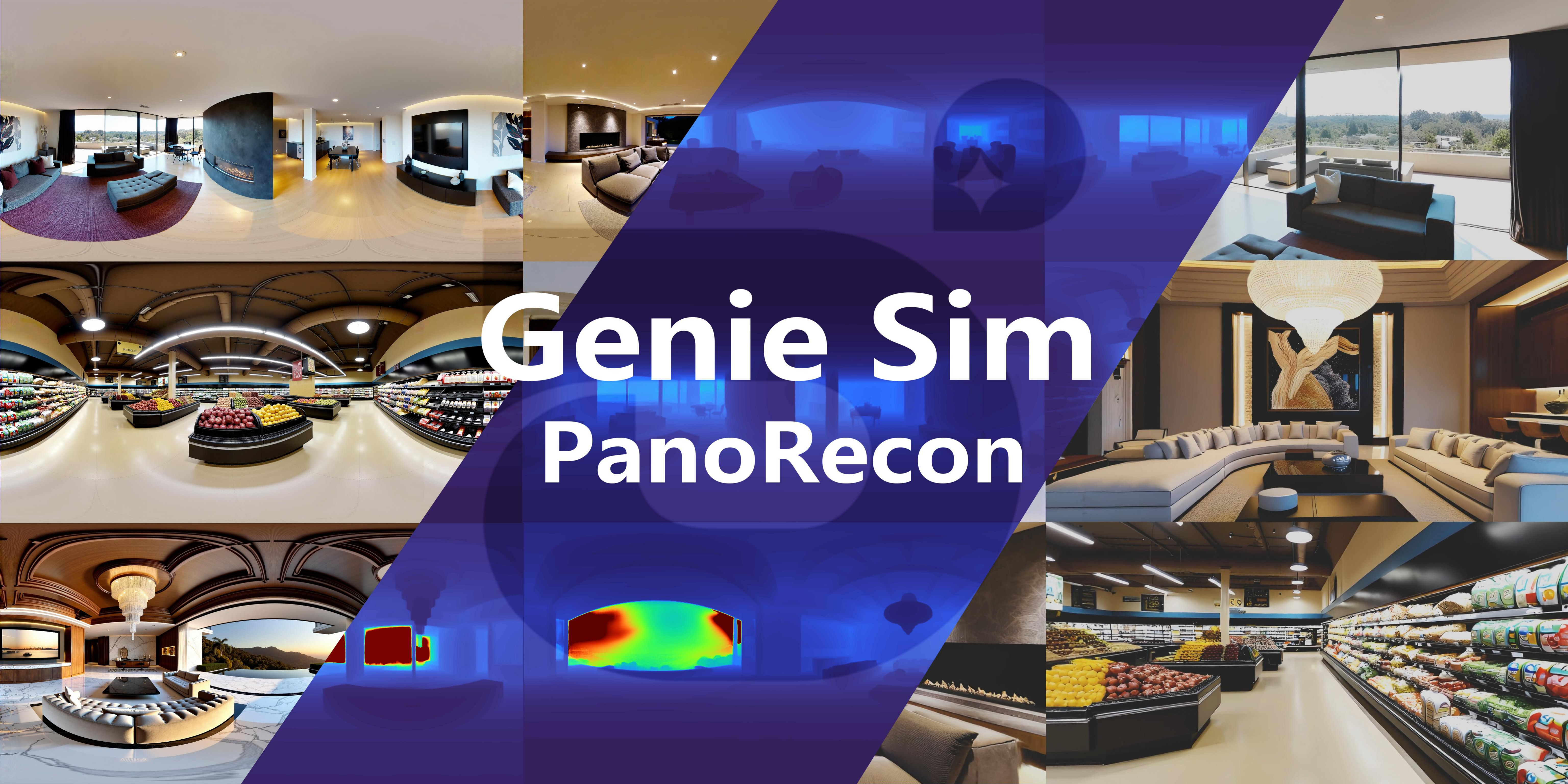} \\[15pt]
    \begin{minipage}{0.9\linewidth} 
        \raggedright 
        \small 
    \end{minipage} 
    \vspace{10pt} 
}
\begin{document}

\maketitle
\pagestyle{empty}


\begin{abstract}
We present Genie Sim PanoRecon, a feed-forward Gaussian-splatting pipeline that delivers high-fidelity, low-cost 3D scenes for robotic manipulation simulation. The panorama input is decomposed into six non-overlapping cube-map faces, processed in parallel, and seamlessly reassembled. To guarantee geometric consistency across views, we devise a depth-aware fusion strategy coupled with a training-free depth-injection module that steers the monocular feed-forward network to generate coherent 3D Gaussians. The whole system reconstructs photo-realistic scenes in seconds and has been integrated into Genie Sim - a LLM-driven simulation platform for embodied synthetic data generation and evaluation — to provide scalable backgrounds for manipulation tasks. For code details, please refer to: \url{https://github.com/AgibotTech/genie_sim/tree/main/source/geniesim_world}.

\end{abstract}

\section{Introduction}
The rapid advancement of research in robotics simulation and embodied intelligence has given rise to an urgent demand for massive high-quality 3D scene data \cite{zhang2025generativeartificialintelligencerobotic}\cite{li2025developmentschallengesdexterousembodied}. For applications ranging from policy training and sim-to-real transfer \cite{dershan2023facilitatingsimtorealintrinsicstochasticity}\cite{aljalbout2025realitygaproboticschallenges} to large-scale synthetic data generation \cite{chen2025robotwin20scalabledata}\cite{gong2026anytaskautomatedtaskdata}, simulation platforms require immersive environments that possess both local photorealism and consistent global spatial structure \cite{geniesim}. Nevertheless, existing scene construction methods still suffer from significant bottlenecks in large-scale production. Iterative optimization approaches represented by 3D Gaussian Splatting (3DGS) \cite{3dgs}\cite{zhu20243dgaussiansplattingrobotics} can achieve high-fidelity reconstruction \cite{mallick2024taming3dgshighqualityradiance}, yet they heavily rely on dense multi-view image inputs, leading to high data acquisition costs. Moreover, they necessitate tedious per-scene optimization for each individual scene, which is time-consuming and exhibits poor generalization ability, making them difficult to meet industrial-level data production requirements \cite{xia2026sagescalableagentic3d}\cite{embodiedgen}.

To improve generation efficiency, a variety of feed-forward 3D reconstruction schemes have emerged in recent years   \cite{pixelsplat}\cite{depthsplat}\cite{anysplat}\cite{zhang2026sparsesplatapplicablefeedforward3d}\cite{ml-sharp}. Such methods dispense with per-scene training and can rapidly generate scene representations via a single forward inference pass. However, existing feed-forward models are mostly designed for conventional pinhole camera images and can hardly be directly adapted to 360$^\circ$ equirectangular inputs \cite{pansplat}\cite{panosplatt3r}. Current practices typically project equirectangular images into multiple independent cube-map faces and perform inference separately \cite{da360}. Nevertheless, independent prediction without global geometric priors tends to cause structural discontinuities and scale drifting at inter-face boundaries, ultimately destroying the spatial continuity of the scene and failing to satisfy the requirements of global geometric consistency for robot navigation and manipulation. 

To address the aforementioned limitations, this paper proposes a panoramic consistent depth-guided feed-forward Gaussian-splatting pipeline. We fuse global structure and local fine-grained depth within the panoramic space to produce a unified geometric constraint map, which drives feed-forward Gaussian generation for individual views, and finally accomplishes scene merging within a few seconds. While maintaining the efficiency of feed-forward inference, the proposed method effectively alleviates the geometric misalignment caused by independent reconstruction across multiple viewports.

This work mainly targets indoor desktop-scale scenes with moderate parallax, aiming to provide a low-cost and scalable solution for generating 3D background assets for robotic manipulation simulation. The pipeline can output geometrically coherent simulation-ready assets without per-scene training, offering an efficient and practical technical pathway for the large-scale construction of data infrastructure for embodied intelligence, and it can be integrated into embodied scene generation toolchains to support scalable manipulation task backgrounds.

\section{Background and Related Work}
\label{sec:related}
3DGS \cite{3dgs} offers high-fidelity, real-time rendering, which led \cite{embodiedgen}\cite{RoboGSim}\cite{SplatSim}\cite{Real-to-Sim} to apply it to embodied simulation to narrow the sim-to-real gap. While platforms like Genie Sim combine foreground meshes with background 3DGS \cite{3dgs} to ensure both physical interactability and visual realism, traditional 3DGS \cite{3dgs} lacks explicit geometry for sim-ready assets. Moreover, its reliance on dense scene-specific capture and time-consuming per-scene optimization severely hinders scalable data production.

To address this, Pixelsplat \cite{pixelsplat}, Depthsplat \cite{depthsplat}, Anysplat \cite{anysplat}, Voxelsplat \cite{voxelsplat} propose feed-forward 3DGS, directly predicting Gaussian parameters via depth estimation networks to bypass lengthy iterative optimization. Methods like PanSplat \cite{pansplat} and PanoSplat3r \cite{panosplatt3r} adapt this to panoramic images but struggle with high-resolution reconstruction and hole-filling due to their pixel-aligned architectures. Conversely, SHARP \cite{ml-sharp} achieves high-definition monocular reconstruction and strong scene generalization using a lightweight Gaussian head, significantly outperforming counterparts in novel view synthesis and unseen region completion.

In our work, we extend SHARP \cite{ml-sharp} to panoramic images via simple, training-free modifications. By substituting traditional 3DGS \cite{3dgs} with feed-forward Gaussians derived from panoramas, we generate realistic backgrounds for embodied manipulation tasks. Our approach rapidly converts generated or captured panoramas into explorable, high-fidelity 3D Gaussian scenes, drastically reducing the production cost of simulation environments.

\section{Method}
\label{sec:method}


\begin{figure}[!t]
\centering
\includegraphics[width=1.0\linewidth]{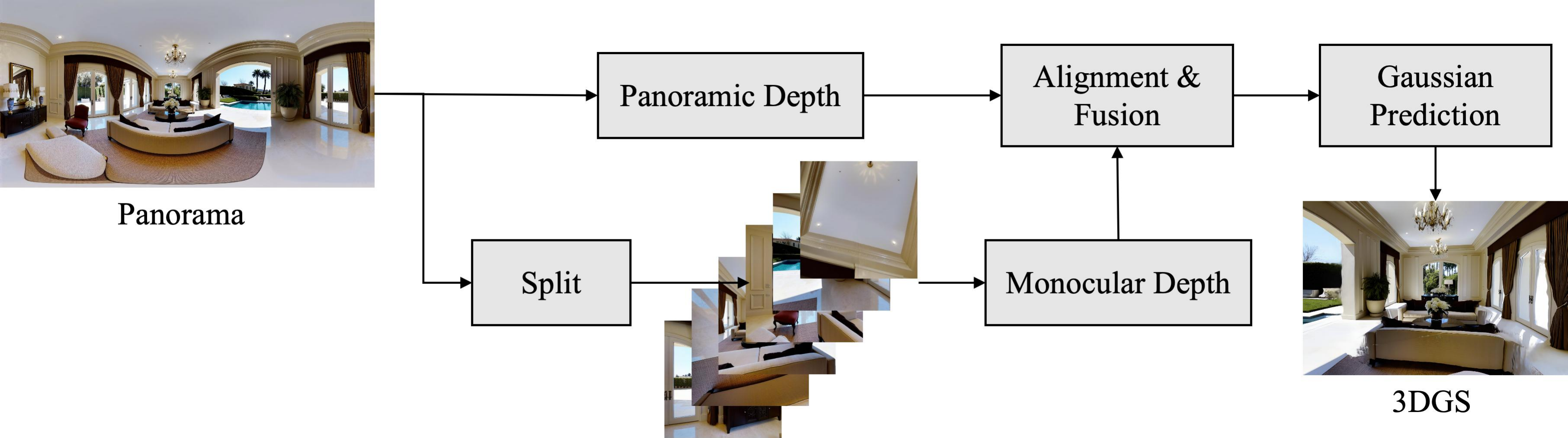} 
\caption{\textbf{Overview of \gsw pipeline.} An input panorama is processed to extract global structural depth (via DA360) and high-resolution local depth details (via DepthPro). These depths are aligned and fused using an inverse-depth Laplacian pyramid. The fused panoramic depth and RGB are then projected into six cubemap faces, serving as geometric constraints to drive a training-free, depth-guided feed-forward Gaussian generation (via SHARP). Finally, the individual faces are merged to form a globally consistent 3D scene.}
\label{fig:pipeline}
\end{figure}

\begin{figure}[hb]
    \centering
    \includegraphics[width=1.0\linewidth]{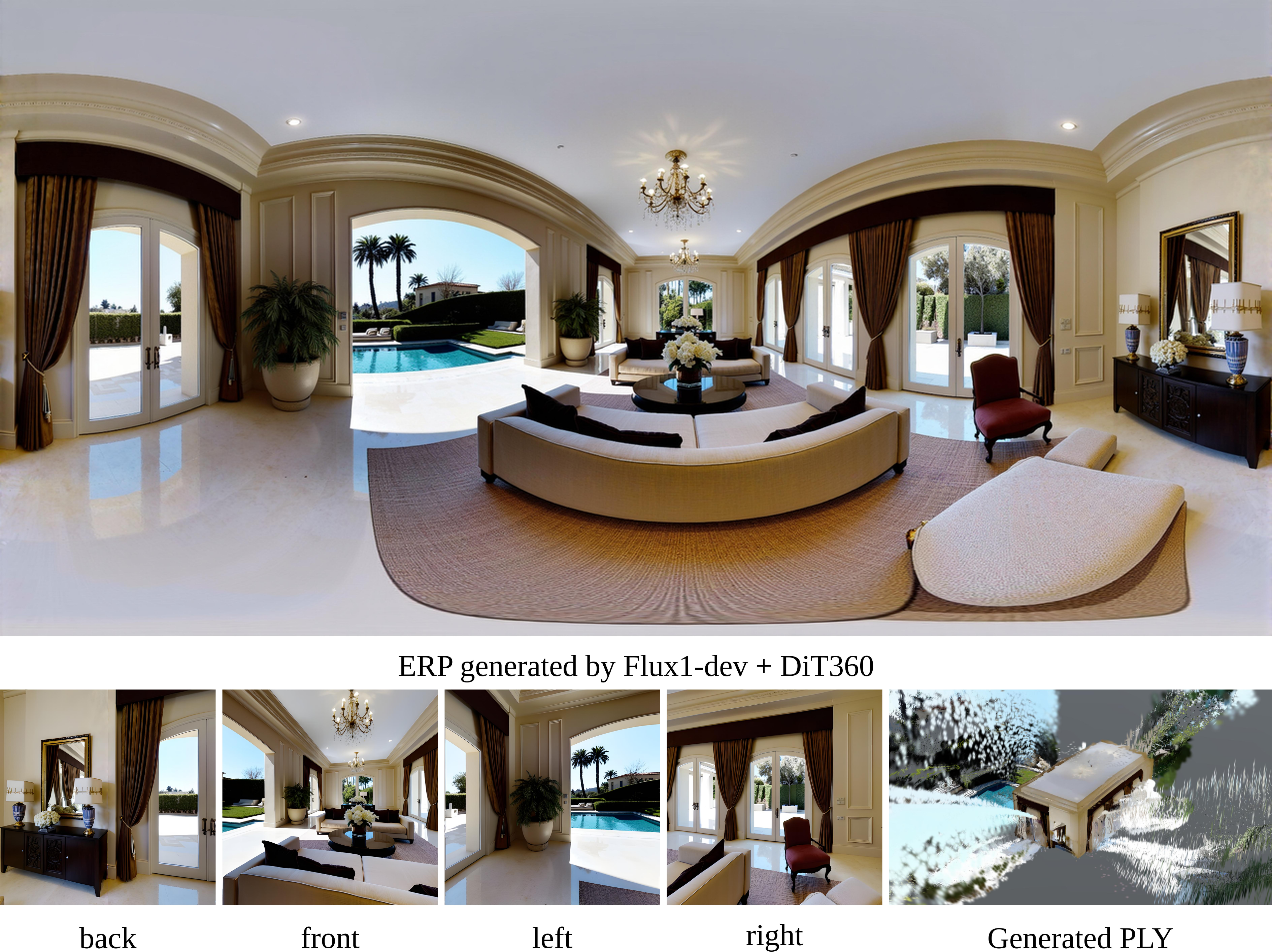}
    \caption{\textbf{PanoRecon Overview.} Top: Input single-view panorama. Bottom: Decomposed cubemap faces and the final reconstructed 3D Gaussian scene.}
    \label{fig:overview}
\end{figure}



\subsection{Panoramic Depth Fusion}
Splitting a panoramic image into six cubemaps for individual feedforward reconstruction preserves high-resolution details, yet often causes inconsistent depth scales and visible seam artifacts between adjacent cubemaps. To address this problem, we present a panoramic depth fusion strategy to generate high-resolution and scale-consistent depth maps. Specifically, we first estimate a panoramic depth map from DA360 \cite{da360}, which is then projected into cubemap format and upsampled to maintain spatial consistency. Meanwhile, we leverage DepthPro \cite{depthpro} to predict a high-resolution monocular depth map that contains rich fine-grained details but suffers from scale inconsistency. Finally, a four-layer Laplacian pyramid is adopted to fuse the low-frequency global scale cues from DA360 inverse depth and the high-frequency texture details from DepthPro \cite{depthpro} inverse depth, as illustrated in \ref{fig:pipeline}.


\subsection{Training-free Depth Injection}
In mainstream feedforward Gaussian reconstruction frameworks, the Gaussian prediction head is cascaded behind a deep depth estimation network, where Gaussian parameters are inferred from both the depth output and latent features of the DPT \cite{dpt} head. In this work, we demonstrate that by substituting the depth map generated by the original depth estimation backbone and revising the corresponding Gaussian initialization scheme, the feedforward pipeline of SHARP \cite{ml-sharp} can directly synthesize geometrically consistent Gaussian scenes under predefined depth constraints in a training-free manner. Accordingly, we replace the intermediate depth predictions in SHARP \cite{ml-sharp} with our fused depth maps to perform feedforward Gaussian reconstruction for each cubemap. Afterwards, we eliminate Gaussian splats lying outside each cuboid frustum and merge all valid Gaussians to construct the complete panoramic 3D Gaussian scene.

\section{Experiments}
\label{sec:exp}

\subsection{Data}
For each experiment, we report the input panorama resolution, including real panoramas (e.g., indoor captures), synthetic panoramas (including DiT360 outputs), whether super-resolution is applied, and the GPU configuration used for inference. The evaluation focuses on scenarios aligned with Genie Sim applications, including diverse indoor environments, table-top manipulation settings, and scalable panorama generation for synthetic data expansion. We emphasize methodological comparisons rather than large-scale benchmark evaluations.

\subsection{Main Comparisons}

\begin{figure}[h]
    \centering
    \begin{tabular}{ccc}

        \includegraphics[width=0.3\linewidth]{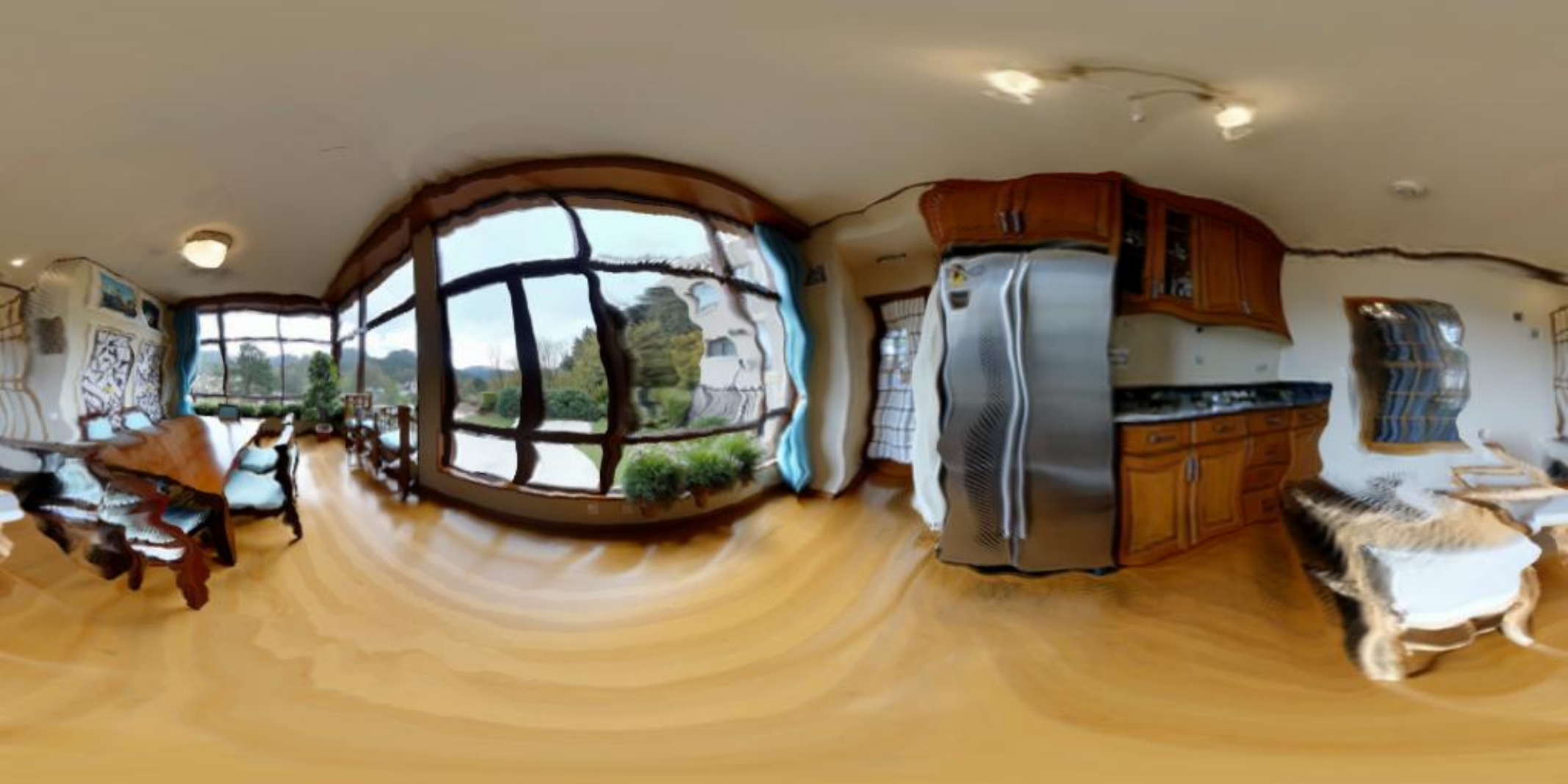} &  
        \includegraphics[width=0.3\linewidth]{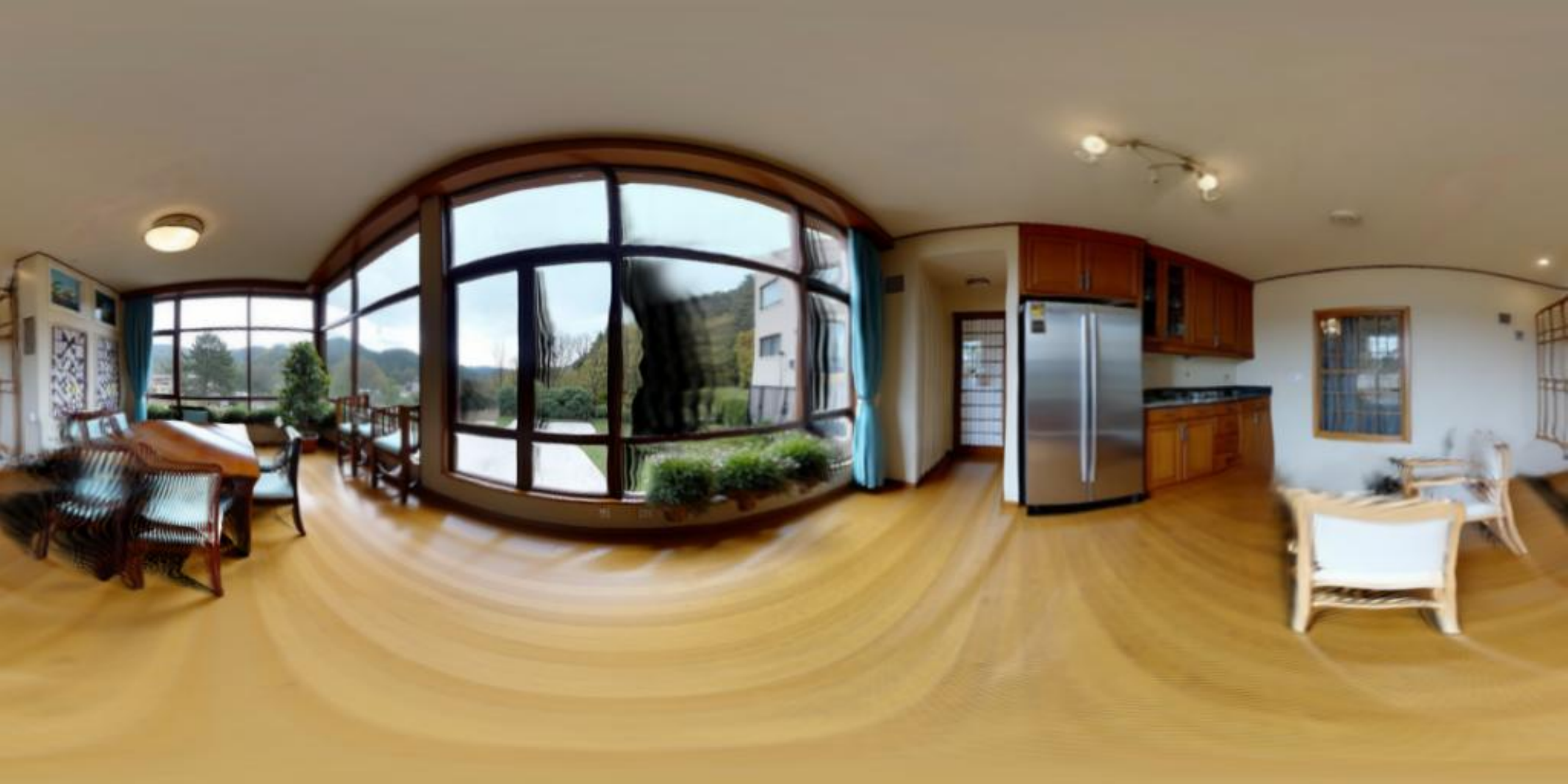} &  
        \includegraphics[width=0.3\linewidth]{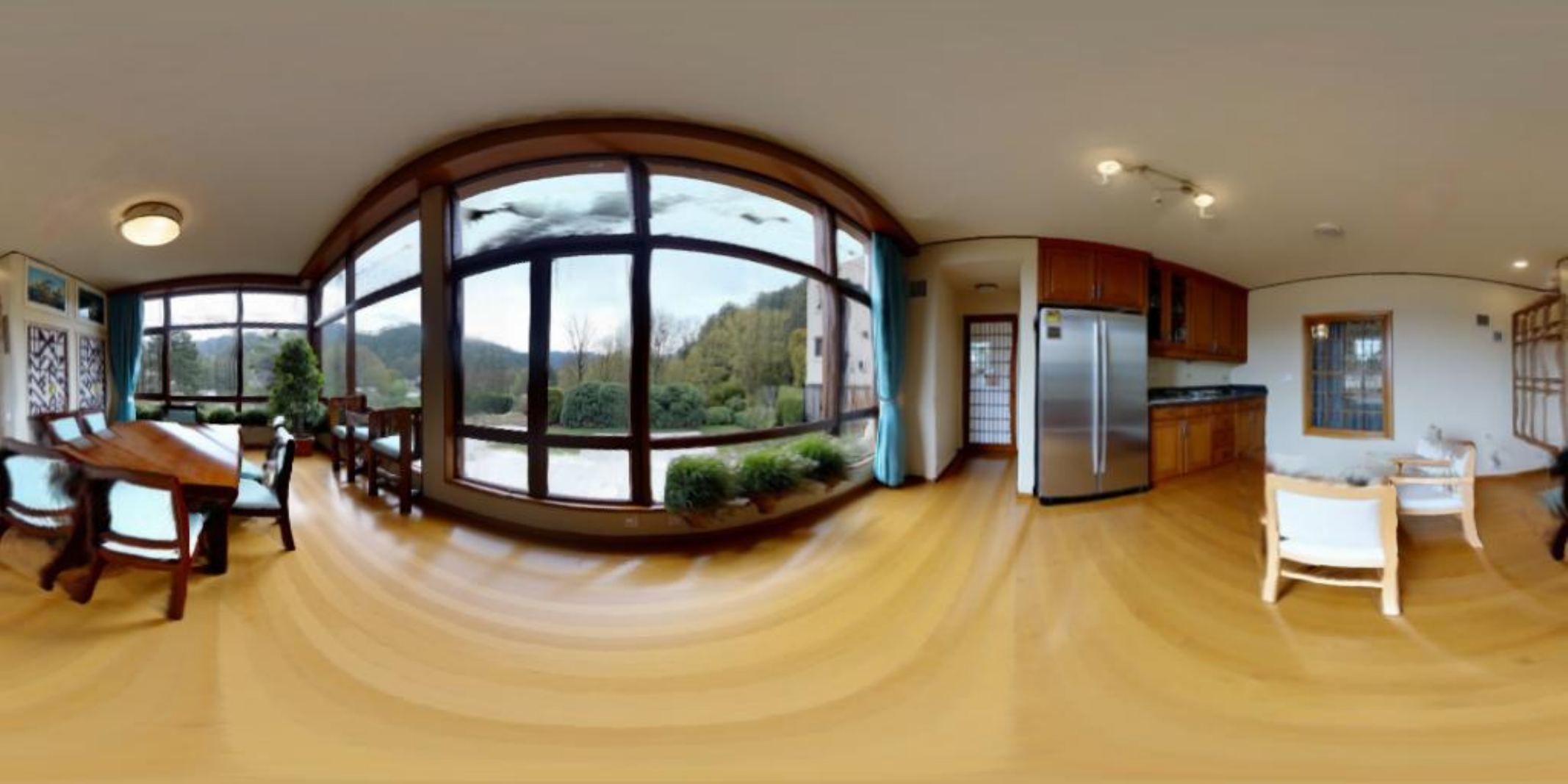} \\
        \includegraphics[width=0.3\linewidth]{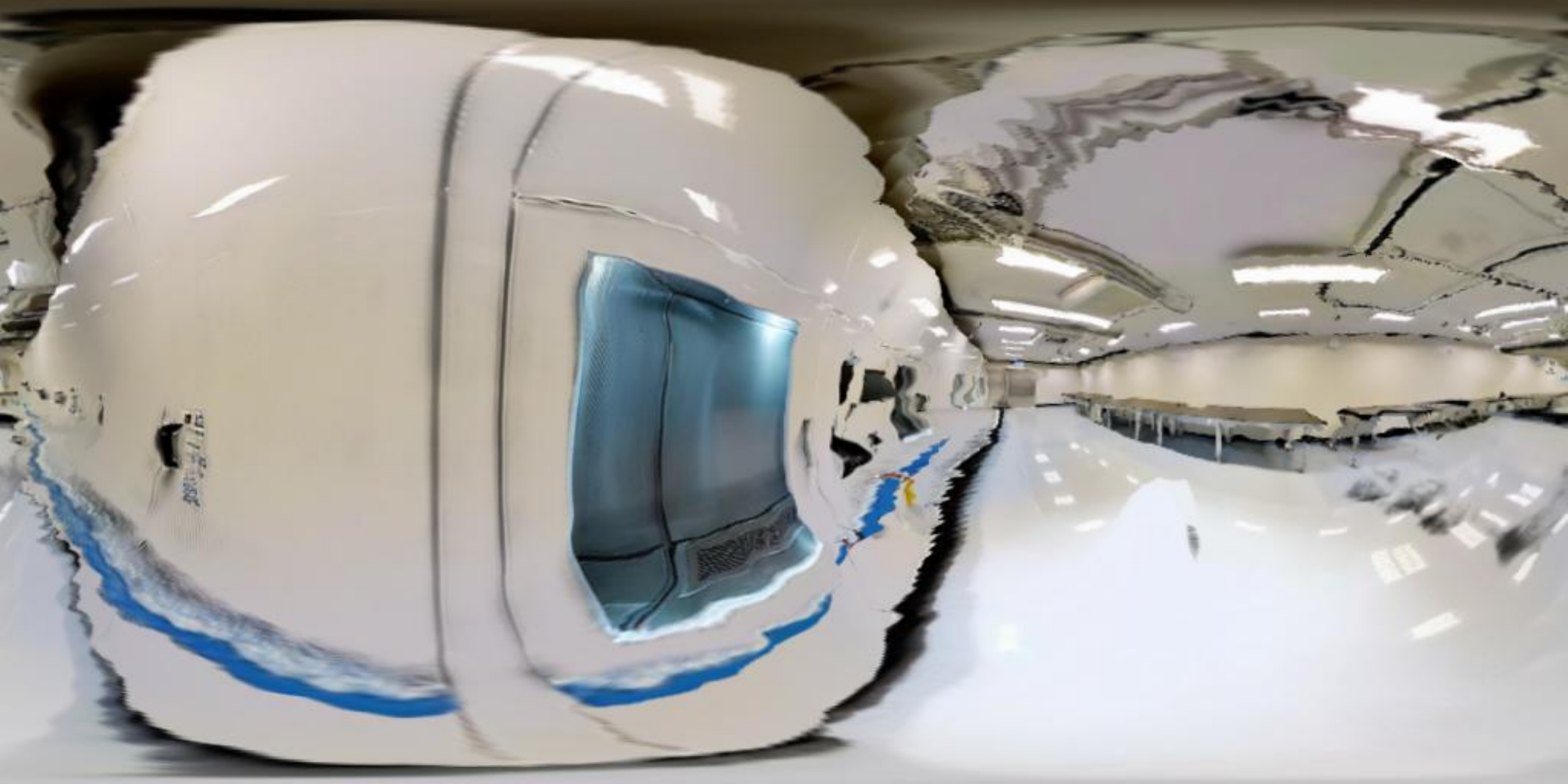} &  
        \includegraphics[width=0.3\linewidth]{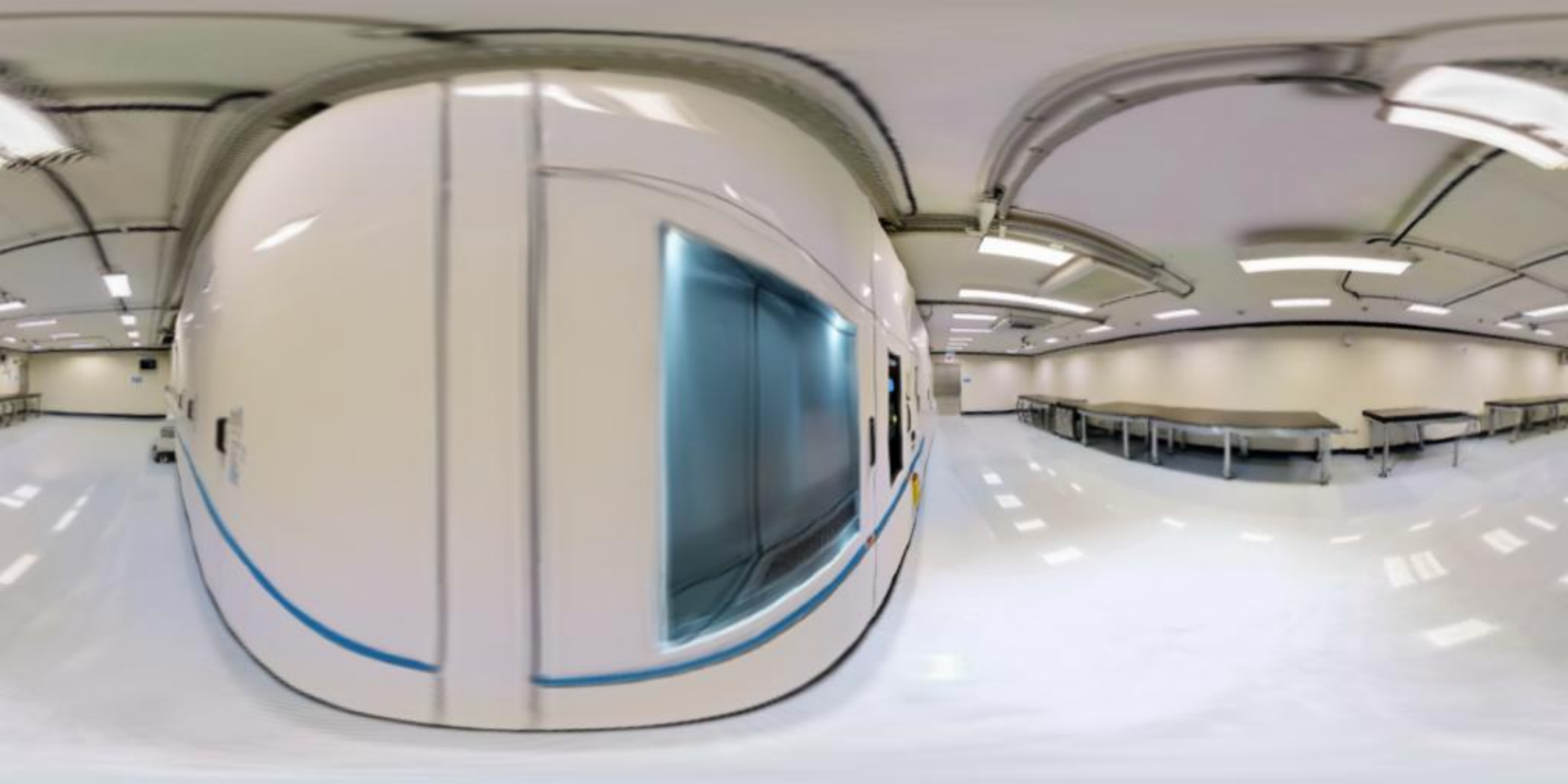} &  
        \includegraphics[width=0.3\linewidth]{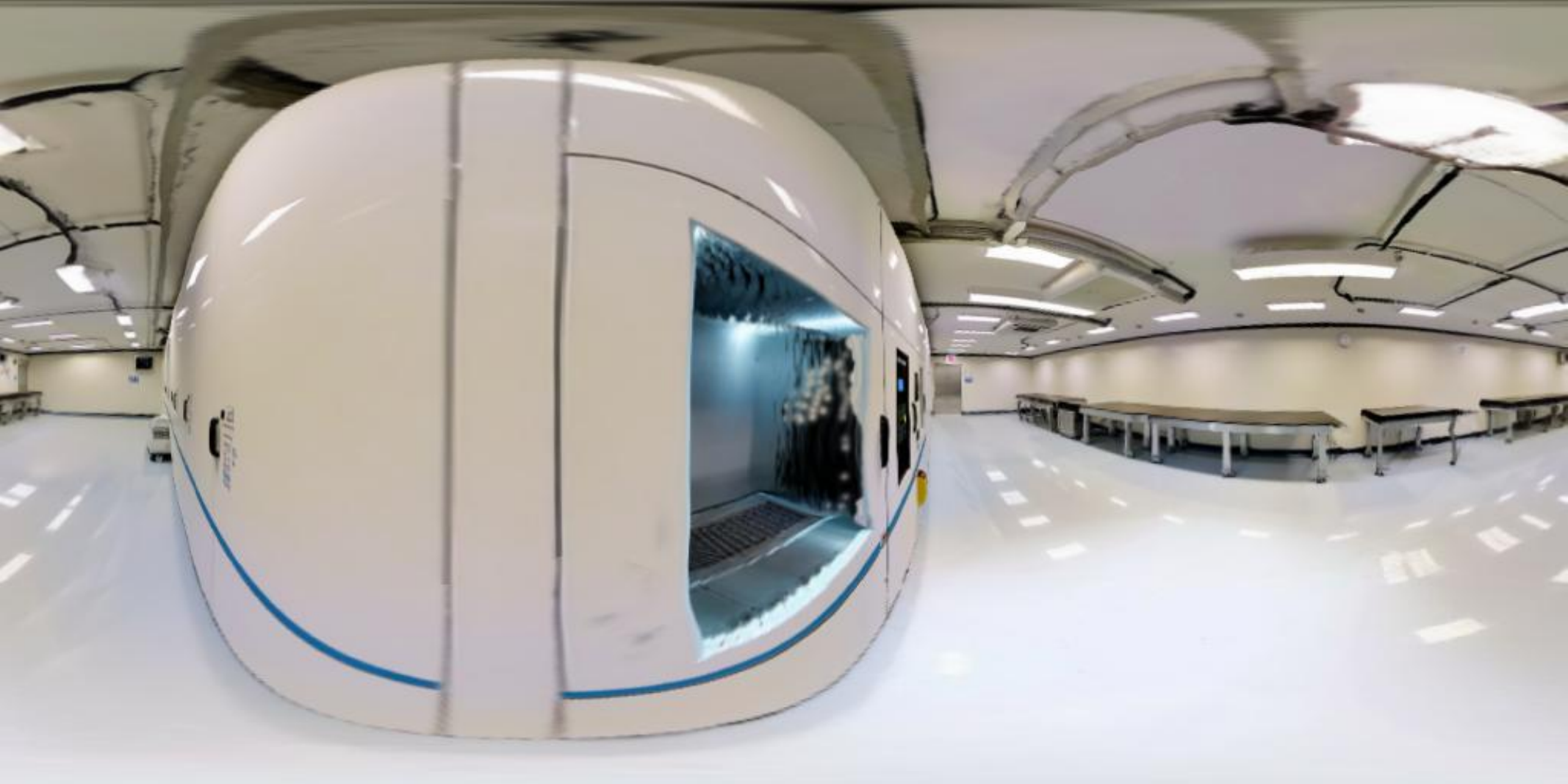} \\
        \includegraphics[width=0.3\linewidth]{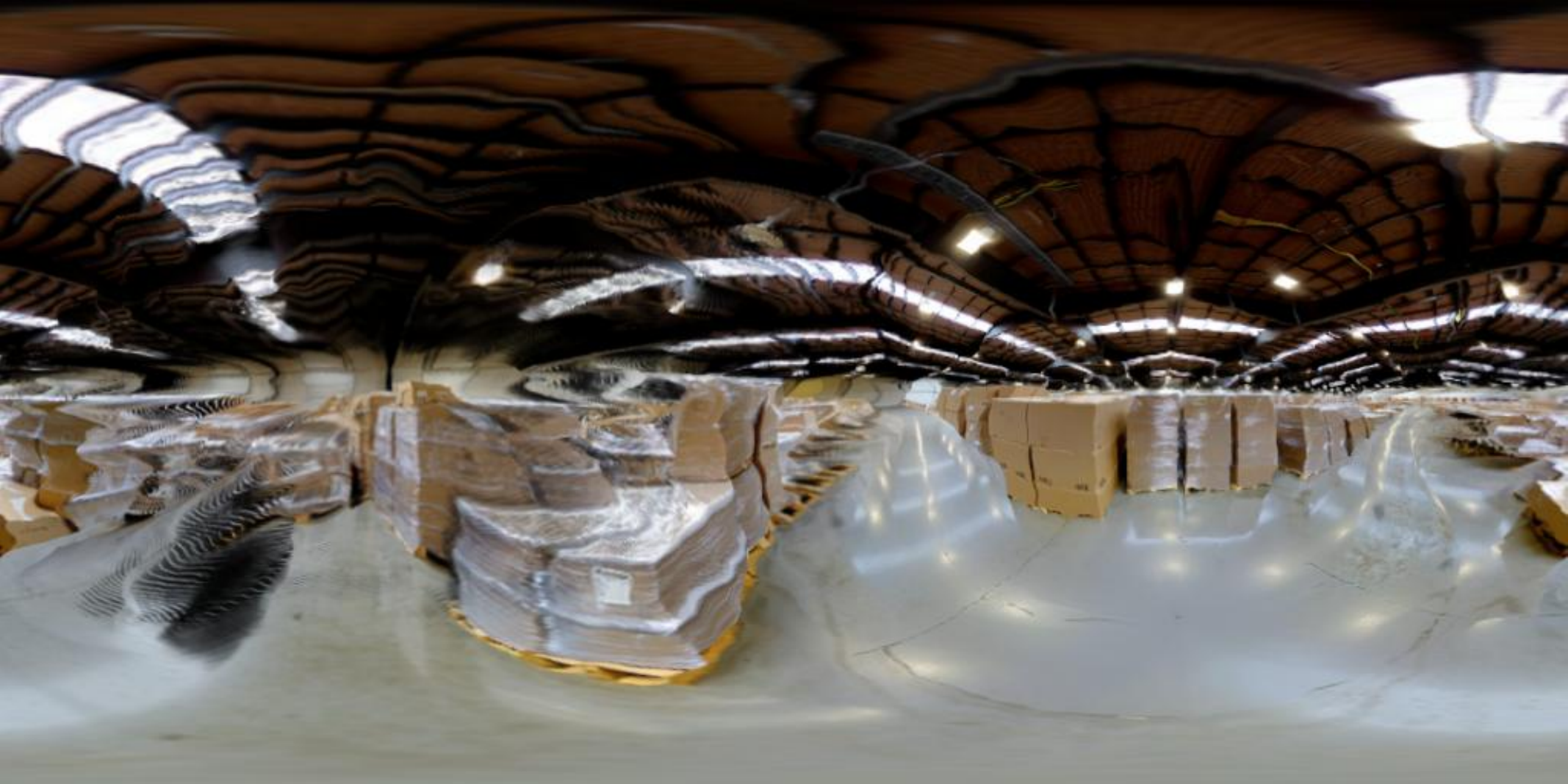} &  
        \includegraphics[width=0.3\linewidth]{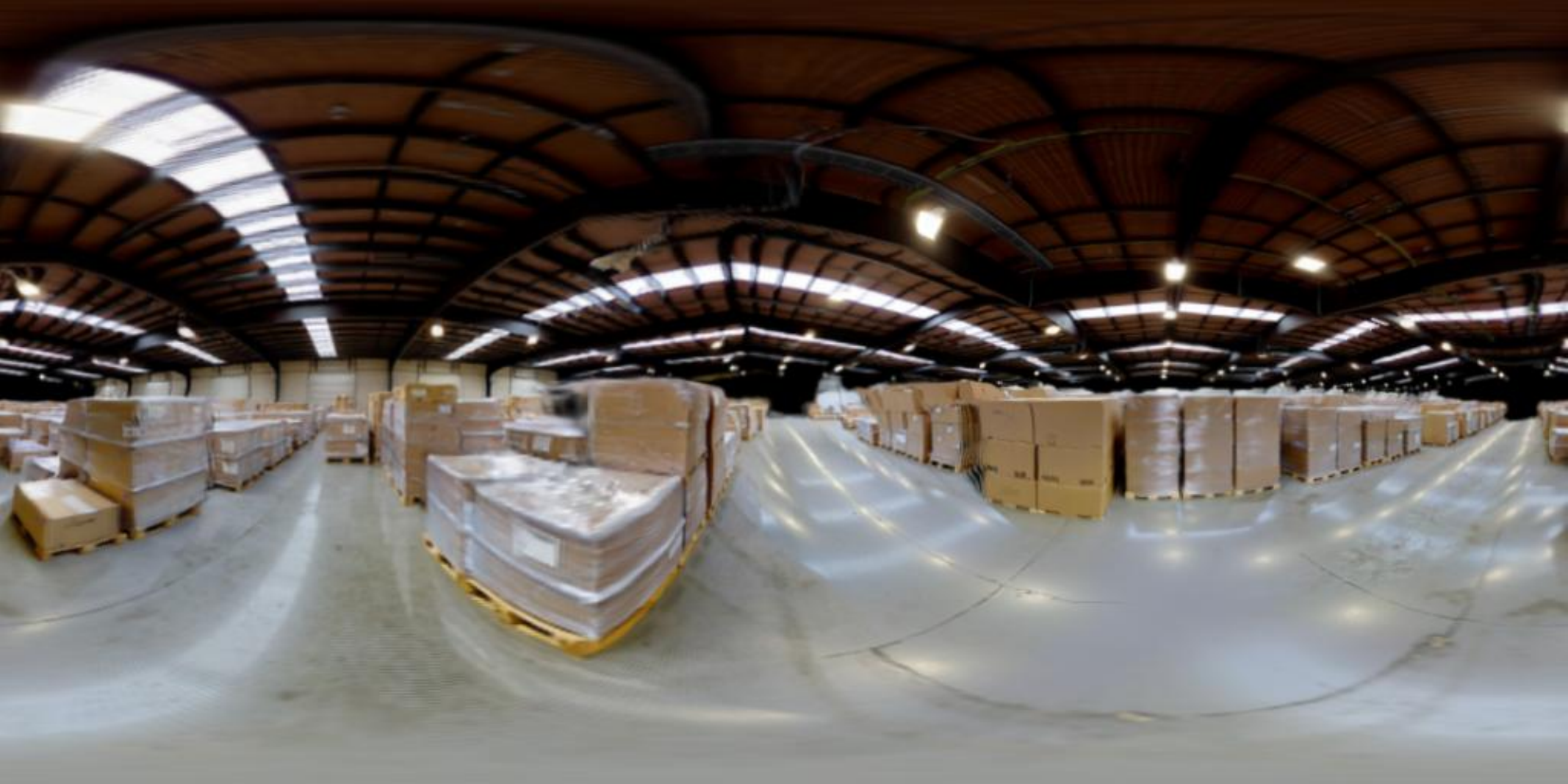} &  
        \includegraphics[width=0.3\linewidth]{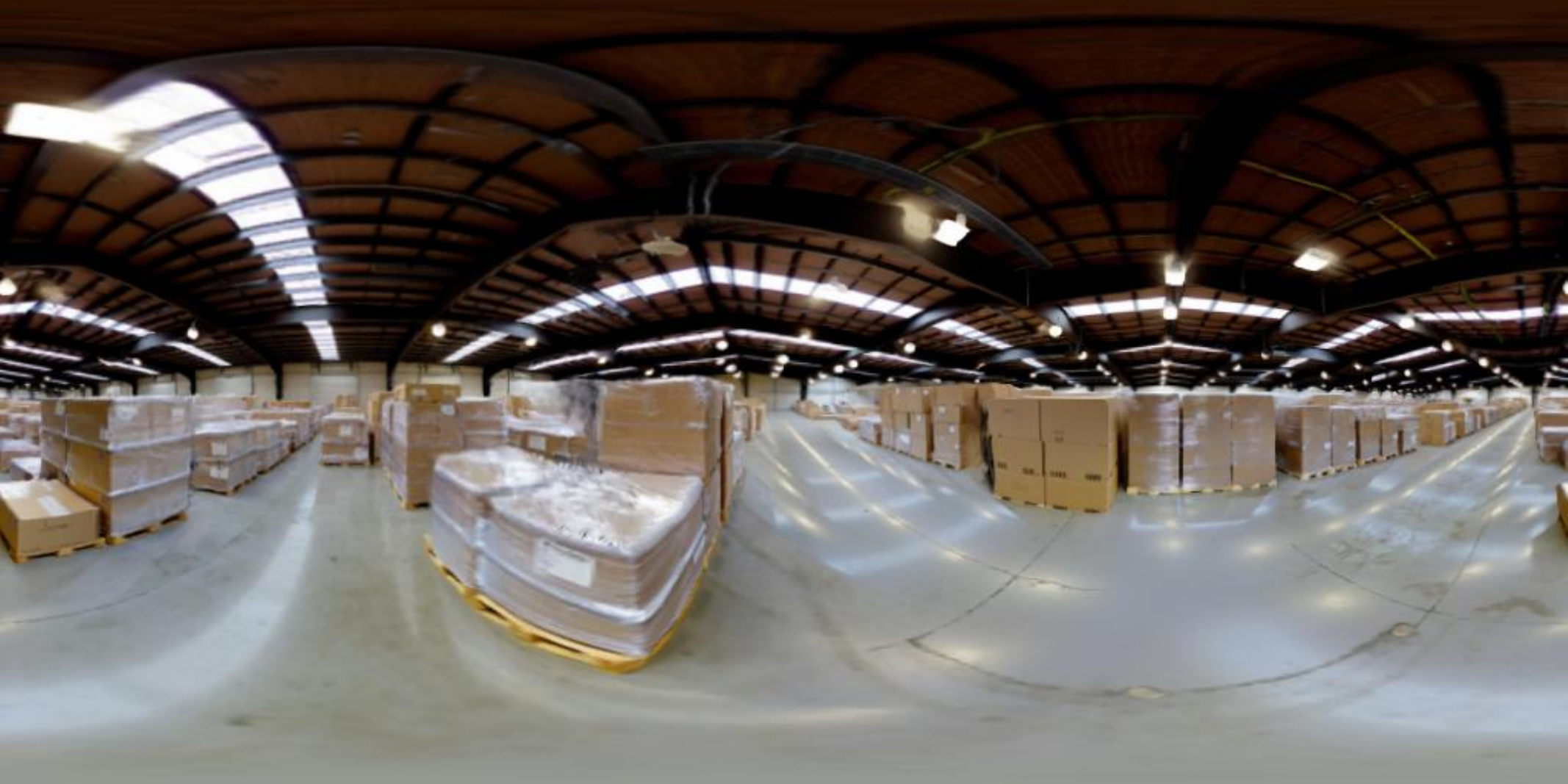} \\
        \textbf{PanSplat}\cite{pansplat} & \textbf{PanoSplatt3R}\cite{panosplatt3r} & 
        \textbf{Ours} \\
        
    \end{tabular}
        \caption{\textbf{Novel View Synthesis Comparison.} }
    
    \label{fig:nvs_comp}
\end{figure}
We conducted a novel view synthesis comparison between PanSplat\cite{pansplat} and PanoSplatt3R\cite{panosplatt3r}, using an identical panoramic pair as input. As shown in Fig. \ref{fig:nvs_comp}. our method yields sharper geometry and more photorealistic appearance in novel views.

\begin{figure}[ht]
    \centering
    \includegraphics[width=1.0\linewidth]{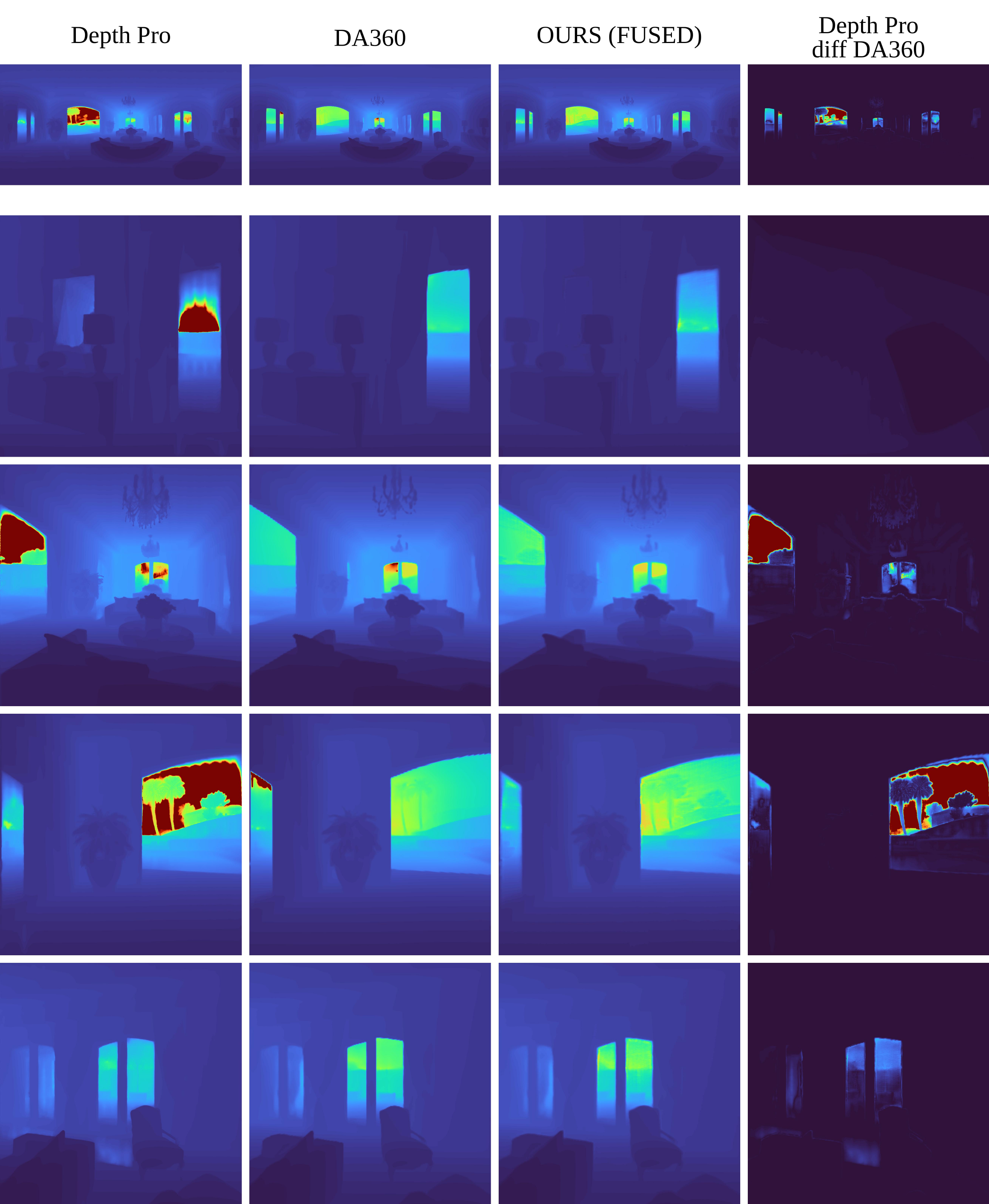}
    \caption{\textbf{Panoramic Depth Comparison.} DA360 preserves global structure but lacks detail, whereas DepthPro offers sharp details with scale inconsistency. Our fusion achieves both accurate global constraints and sharp local boundaries.}
    \label{fig:Depth_Pano_Comparision}
\end{figure}

We also compare different system modes that vary in how depth is obtained and whether cross-view consistency is enforced, and the results are shown in Fig. \ref{fig:Depth_Pano_Comparision} .



\begin{figure}[ht]
    \centering
    \includegraphics[width=1.0\linewidth]{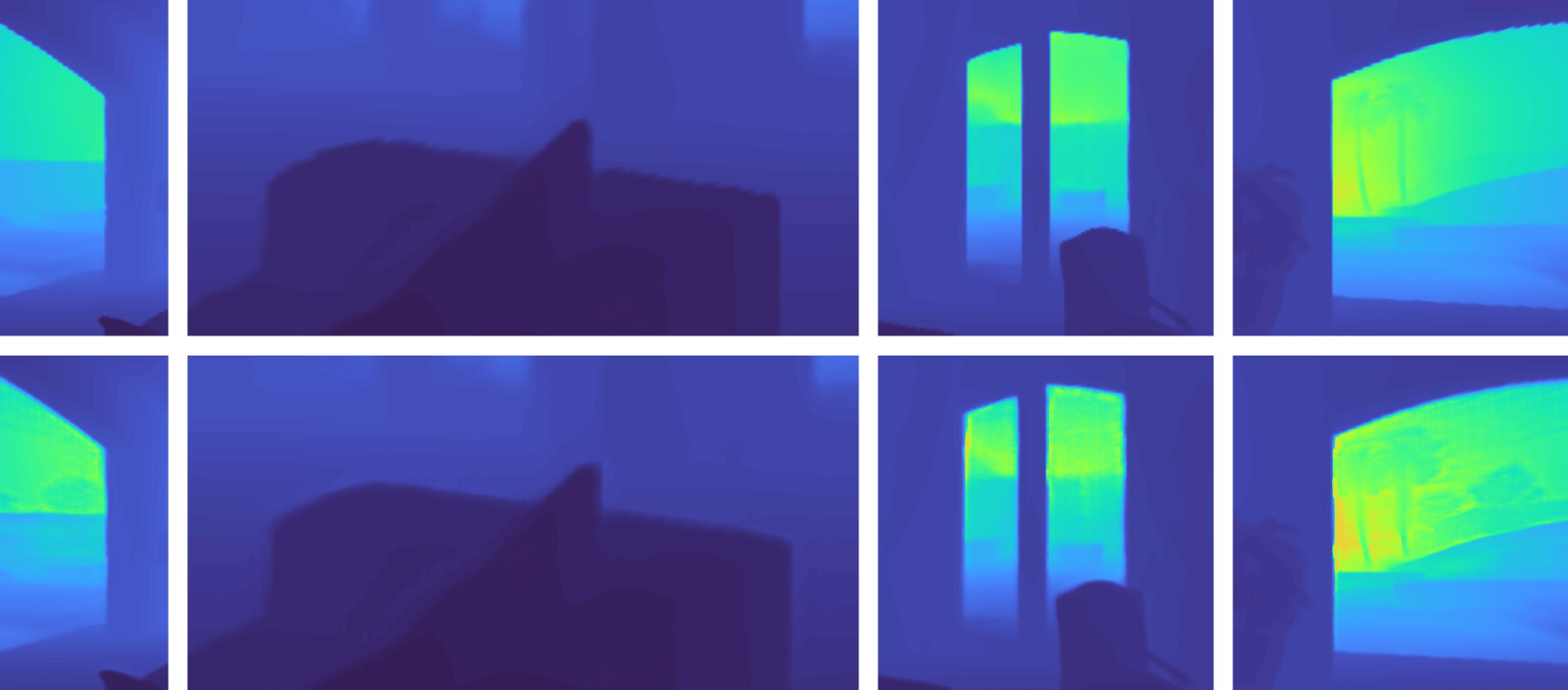}
    \caption{\textbf{Effect of Anti-aliasing.} Applying anti-aliasing during cubemap projection mitigates stair-step artifacts and boundary noise, enabling accurate feed-forward Gaussian initialization.}
    \label{fig:anti_aliasing_combined}
\end{figure}

\begin{figure}[ht]
    \centering
    \includegraphics[width=1.0\linewidth]{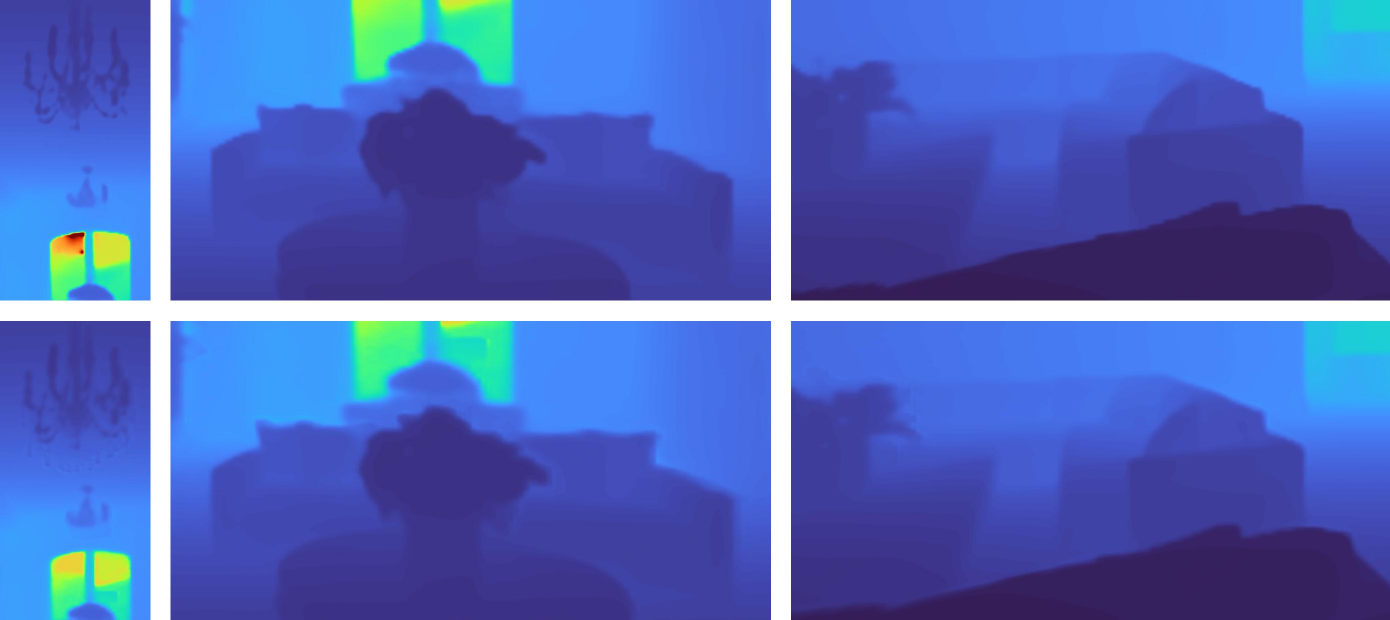}
    \caption{\textbf{Preservation of Fine-Grained Details.} By integrating high-frequency depth cues, our pipeline effectively captures intricate local structures. Compared to relying solely on global depth priors, our proposed method yields significantly sharper object boundaries and preserves tiny geometric details in the reconstructed 3D scene.}
    \label{fig:more_details_combined}
\end{figure}

\begin{figure}[ht]
    \centering
    \includegraphics[width=1.0\linewidth]{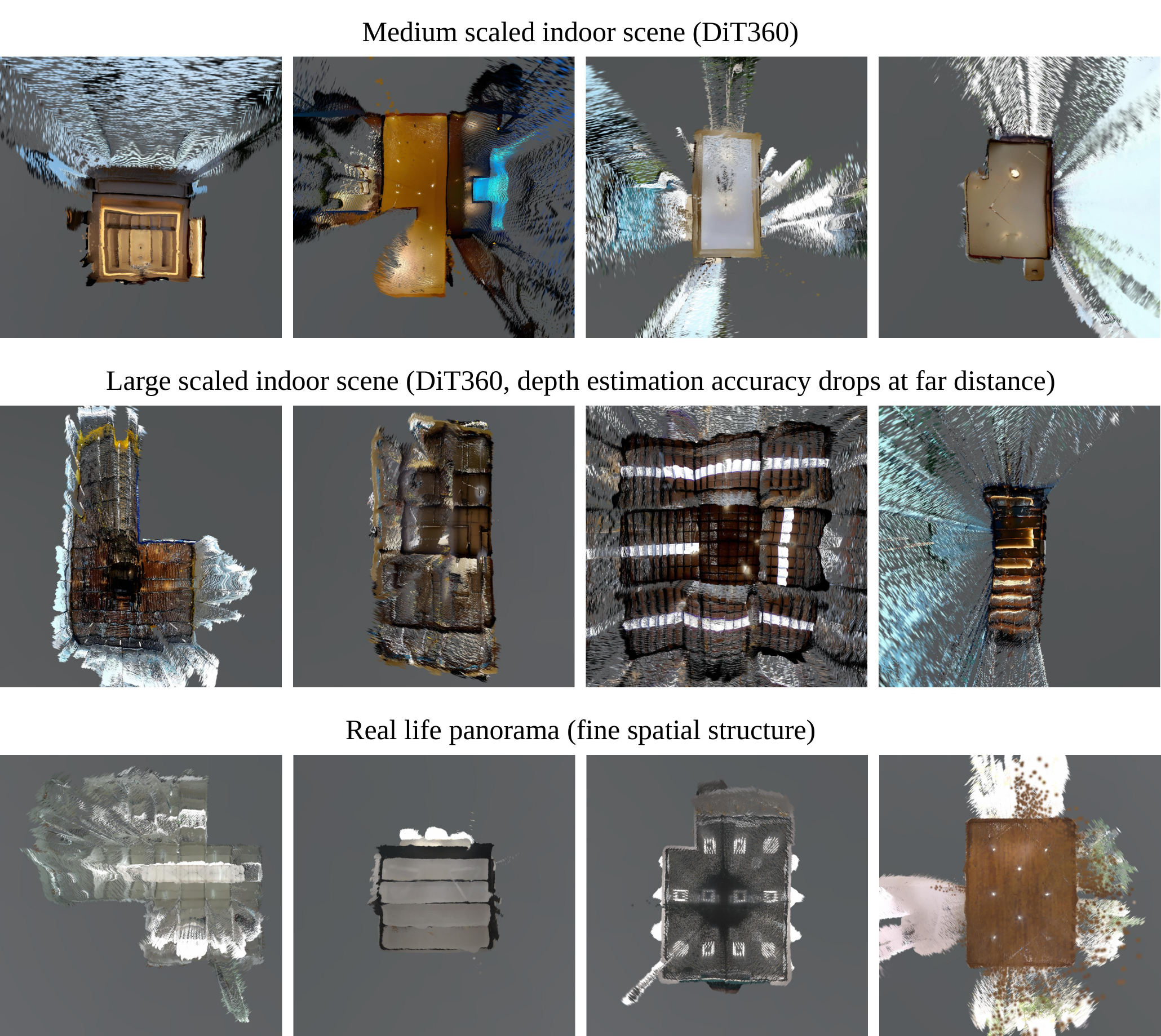}
    \caption{\textbf{Accurate Spatial Geometry.} The reconstructed 3D Gaussian scenes exhibit highly coherent spatial structures and accurate geometric layouts derived from single-view panoramas (e.g., DiT360 generated inputs). The preserved global consistency and minimal structural distortion make these scenes highly suitable for serving as robust background assets in indoor robotic manipulation simulations.}
    \label{fig:spatial_geometry_combined}
\end{figure}



\subsection{Metrics and Reporting}

Qualitative evaluation (primary): We assess seam consistency across independently reconstructed views, stability in polar regions, and geometric coherence of the merged Gaussian scene under novel view rendering.

Quantitative evaluation: We report wall-clock runtime for each pipeline stage and peak GPU memory usage. Additionally, we optionally evaluate depth consistency across view boundaries and include user studies when applicable.


\section{Limitations and Future Work}
\label{sec:limits}


Our current work is based on training-free modifications to SHARP \cite{ml-sharp}. While it achieves decent rendering quality, it still suffers from the inherent limitations of the pixel-aligned feedforward 3D Gaussian method, including the inability to complete unseen regions in novel views. Furthermore, our scope assumes an approximately ${\sim}1$,m$^3$ manipulation space with limited parallax; these are design choices rather than universal conditions for outdoor reconstruction. Finally, any claims regarding deployment to formats such as USDZ and Isaac Sim should be presented as a future outlook rather than validated achievements, unless end-to-end performance has been rigorously measured.

\section{Conclusion}
\label{sec:conc}

We described \gsw, a feedforward pipeline that combines DA360's globally coherent panoramic depth with SHARP's metric panoramic detail via inverse-depth Laplacian fusion, then depth-guided SHARP on cubemap faces and merge. This design targets speed and panorama-level geometric consistency without per-scene optimization---aiming at a strong cost--performance point for immersive scene generation from single panoramic image. Together with the broader \gs simulation and synthetic-data stack~\cite{geniesim}, it supports the same team-level goal: lowering the cost of high-quality, spatially grounded training and evaluation environments while this manuscript's primary contribution stays method-centric.

\section*{Acknowledgment}
Address the Genie Sim team. The \gs platform provides the larger simulation, dataset, and evaluation context. Thank collaborators and third-party maintainers of DA360, SHARP, DiT360, ComfyUI, Real-ESRGAN, and other dependencies.





\printbibliography

\appendix
\section{Supplement / Code}

Implementation and CLI options are documented in the Genie  Sim World codebase. The open \gs repository describes the full simulation platform, synthetic data, and related tooling; cite or link it when positioning this work within AgiBot's embodied-AI ecosystem.





\begin{figure}[ht]
    \centering
    \includegraphics[width=1.0\linewidth]{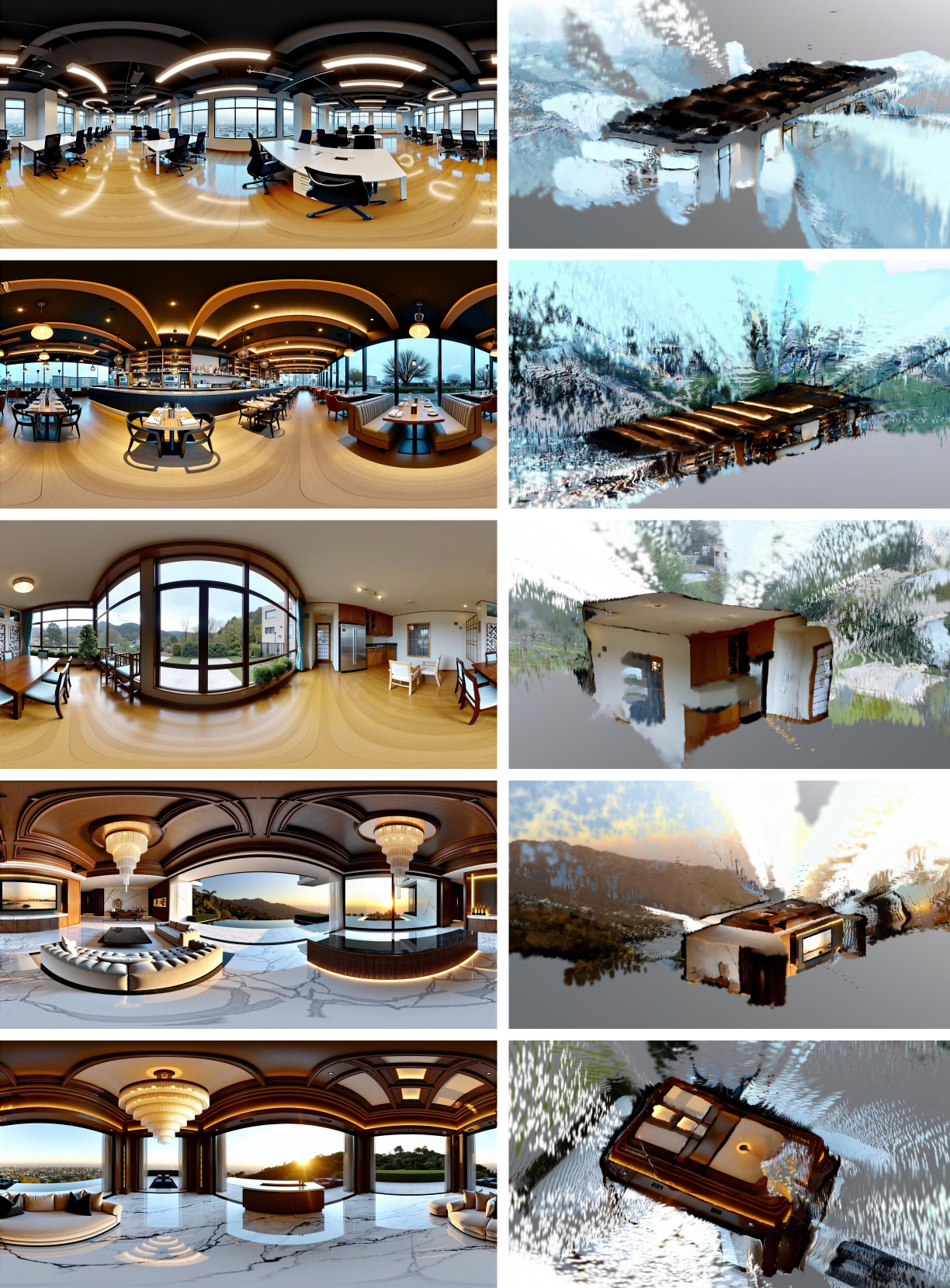}
    \caption{\textbf{More World.}}
    \label{fig:more_world}
\end{figure}
\begin{figure*}[ht]
    \centering
    \includegraphics[width=1.0\linewidth]{appendix_combined_grid.pdf}
    \caption{\textbf{More World.}}
    \label{fig:appendix}
\end{figure*}
\end{document}